# A novel method of predictive collision risk area estimation for proactive pedestrian accident prevention system in urban surveillance infrastructure

**Byeongjoon Noh[1] (powernoh@kaist.ac.kr) and Hwasoo Yeo[1]\* (hwasoo@kaist.ac.kr)**

[1]: Department of Civil and Environmental Engineering, Korea Advanced Institute of Science and Technology, 291 Daehak-ro, Yuseung-gu, Daejeon, South Korea

\**corresponding author*

**Abstract**

Road traffic accidents, especially vehicle-pedestrian collisions in crosswalk, globally pose a severe threat to human lives and have become a leading cause of premature deaths. In order to protect such vulnerable road users from collisions, it is necessary to recognize possible conflict in advance and warn to road users, not post-facto. A breakthrough for proactively preventing pedestrian collisions is to recognize pedestrian's potential risks based on vision sensors such as CCTVs. In this study, we propose a predictive collision risk area estimation system at unsignalized crosswalks. The proposed system applied trajectories of vehicles and pedestrians from video footage after preprocessing, and then predicted their trajectories by using deep LSTM networks. With use of predicted trajectories, this system can infer collision risk areas statistically, further severity of levels is divided as danger, warning, and relative safe. In order to validate the feasibility and applicability of the proposed system, we applied it and assess the severity of potential risks in two unsignalized spots in Osan city, Korea.

# 1. Introduction

Despite advances in traffic safety technologies, road traffic accidents globally still pose a severe threat to human lives and have become a leading cause of premature deaths [1]. Every year, approximately 1.2 million people are killed and 50 million injured in traffic accidents [2], [3]. Vulnerable road users (VRUs), especially pedestrians, are exposed to various hazards, like drivers failing to yield to them in crosswalks [2]. According to international institutes such as British Transport and Road Research Laboratory and World Health Organization (WHO), crossing roads at unsignalized crosswalks is as dangerous for pedestrians as crossing roads without crosswalks or traffic signals [4].

In order to protect such VRUs from traffic accident, there are two approaches: (1) centralized processing; and (2) on-field processing, as illustrated in **Figure 1**. The centralized processing approach often manages a large amount of data from field equipment (e.g., closed-circuit television (CCTV), traffic detector, etc.). It monitors and responds to abnormal situations (e.g., jaywalking, collisions, etc.) in overall urban. A typical example is 24-hour CCTV surveillance centers in administrative districts. They analyze historical accidents and their factors [5], [6], and then decide proper locations for retrofitting safety facilities in order to improve the safety in the urban environment. However, the fundamental solution is to be aware of the occurrence of the possible accident in advance and warn to road users, not post-facto. In this respect, the on-field processing approach can respond immediately to abnormal situations (e.g., near-miss collision, etc.) in the field such as a cooperative-intelligent transportation system (C-ITS), which is grafted information communication technology (ICT) onto the existing traffic system. C-ITS can provide some on-field response services; the field equipment can detect road objects, recognize abnormal and risky situations, and warn the target objects from danger. In this study, we apply this approach to preventing vehicle-pedestrian collisions, immediately and proactively.

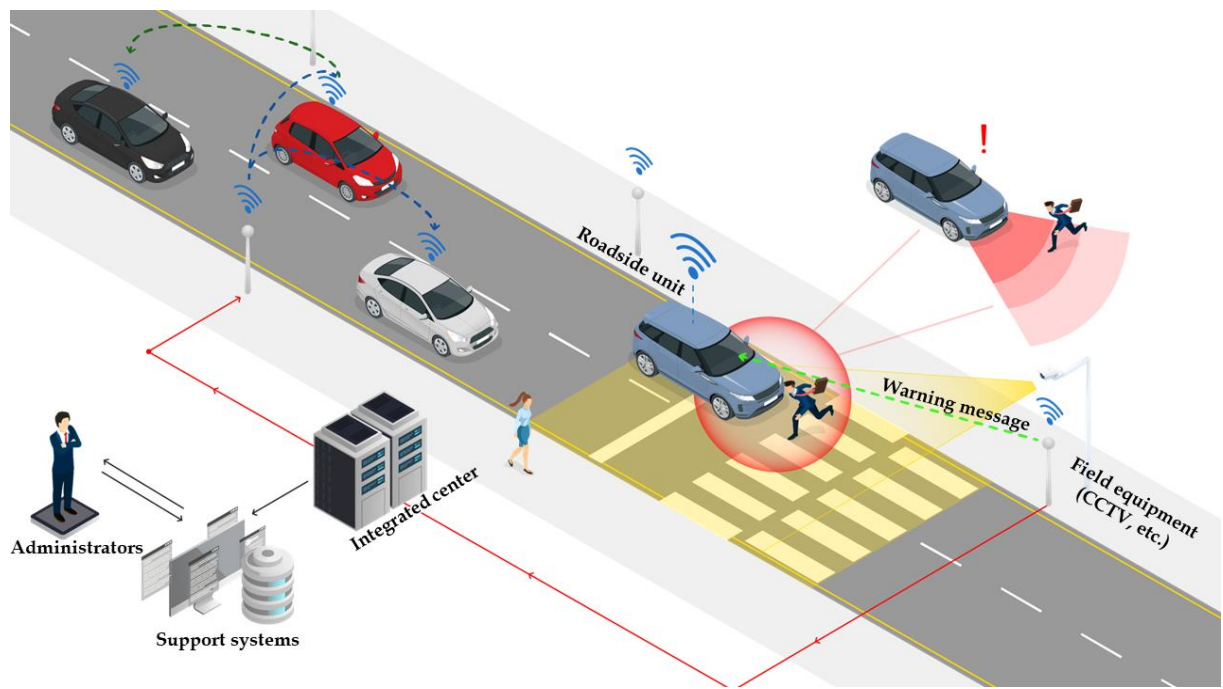

**Figure 1.** Overall approaches to protect VRUs from traffic incidents

A breakthrough for proactively preventing pedestrian collisions is to recognize pedestrian's potential risks (e.g., near-miss collision) based on vision sensors such as CCTVs, which can be applied as field equipment. The use of vision sensors is supposed to make it easier to recognize vehicle-pedestrian collisions in advance and warn immediately them of such situations, and to evaluate vehicle-pedestrian behavioral interactions that pose a threat to pedestrian at crosswalk [7]–[10]. To date, an extensive variety of studies have reported on deriving a surrogate safety measurement (SSM) [11], [12], estimating collision probability [13], measuring levels of potential risks, analyzing road users' behaviors [14], [15] in order to prevent vehicle-pedestrian collisions proactively.

Most of them used the trajectories of road users and the predicted information to achieve their objectives. For example, the authors in [11] analyzed vehicle-pedestrian interactions with speed, time-to-collision (TTC), and gap time (GT), and classified severity levels of their interactions by using Import Vector Machine (IVM). This will be applied as SSM, and support to evaluate various infrastructure and control improvements for making urban interactions safe for road users. The authors in [12] proposed an alternative approach that makes use of traffic conflicts extracted from traffic video recording for safety assessment due to a lack of historical crash data. They used vehicle trajectories as indicators of safety performance, and TTC was applied as conflict risk indicator with TTC* threshold to decide risk related with the perception, reaction time and driving conditions. The authors in [13] estimated vehicle-pedestrian collision probability at intersections. They defined the critical time depending on collision patterns of perception-reaction failure and evasive action failure to identify the latent collision risk for computing collision probability. The authors in [14] predicted pedestrians' red-light crossing intention in a crosswalk by using video data from real traffic scenes. Similar to preprocessing step of our study, they used deep learning-based object detection model (YOLO v3 in [14]) for detecting pedestrians, extracted features manually such as gender and walking direction. Then, with categorizing pedestrian's behaviors as three stages (show up, show intention, start to cross) and defining "crossing intention" pattern, the crossing intentions were predicted by using LSTM-RNN network. The authors in [15] proposed pedestrian path prediction system at a time horizon of 2s by applying waiting/crossing decision model (W/CDM) and modified social force model (MSFM). This can recognize a possible conflict between straight-going vehicles and pedestrians at unsignalized crosswalk, and guide to decide in advance whether autonomous vehicle continue to move forward or stop. These studies are expected to highly improve pedestrian safety in proactive perspective.

Unlike the existing approaches, this study proposes a predictive collision risk area (PCRA) estimation system in order to alleviate vehicle-pedestrian collisions at unsignalized crosswalks. The proposed system applies trajectories of vehicles and pedestrians from video footage after preprocessing, and then predicts their trajectories by using deep LSTM networks. With use of predicted trajectories, this system infers collision risk areas statistically, further, severity of levels is categorized as danger, warning, and relative safe. The main objectives of this study are: (1) to estimate the predictive potential risk areas for vehicle and pedestrian; (2) to measure the potential risks in actual vehicle-pedestrian interactions; and (3) to proactively provide predictive collision warning to driver and pedestrian as a part of C-ITS services.

The remainder of this paper consists of three chapters as follows:

1. Materials and methods: Descriptions of a predictive collision risk area estimation system based on deep LSTM networks and statistical inference method.

2. Experiments and results: Validating feasibility and applicability of the proposed system, and discussion of results and limitations.

3. Conclusions: Summary of our study and future research directions.

## 2. Materials and Methods

2.1. Overall architecture of predictive collision risk area estimation system

In this section, we describe an overall architecture of the proposed system for proactive pedestrian accident prevention employed on real traffic video footage from CCTVs deployed on the road. The proposed system can predict trajectories of vehicles and pedestrians, estimate the predictive collision risk areas for their interactions, and figure out severity of potential risks by levels depending on time second. **Figure 2** illustrates the overall structure of the proposed system, consisting of four parts: 1) data sources; 2) preprocessing; 3) predictive risk area estimation; and 4) multi-dimensional analysis.

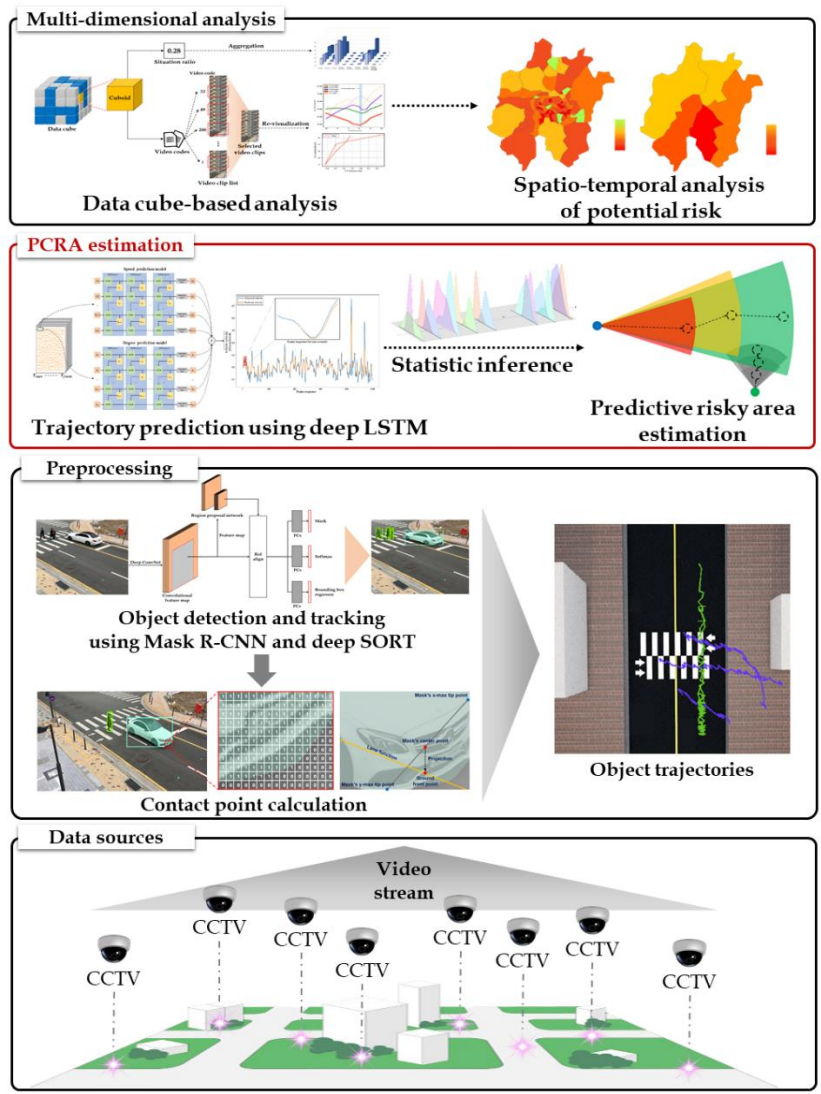

**Figure 2.** Overall structure of the proposed system and PCRA estimation part (redbox)

In the first and second parts, the video stream from CCTVs are handled such as detecting vehicles and pedestrians, tracking them, calculating their contact points, and extracting their trajectories. Mask

R-CNN (Region based Convolutional Neural Network) model [16] and deep SORT (Simple Online and Realtime Tracking) method [17] are used to detect objects and trace them in frame units, respectively. Then, their "contact points" are calculated, and trajectories are extracted. Since this study is an extension of our previous studies, [18]–[21], the detailed descriptions about some methods for data collection and preprocessing parts are omitted. Please refer to our previous studies for the detailed explanation [18]–[20].

In this paper, we aim to describe the third part, trajectory prediction and risk area estimation (redbox in **Figure 2**). This part mainly consists of two modules: 1) object's trajectory prediction using deep LSTM networks; and 2) predictive collision risk area (PCRA) estimation using statistical inference. In the first module, we predict trajectories of vehicles and pedestrians for after a few seconds by applying deep LSTM networks trained by the extracted trajectories. With the predicted trajectories, we estimate PCRA by using statistical inference, which enables to measure severity of potential risks between vehicle and pedestrian.

2.2. Data sources and preprocessing

In our experiments, we used datasets based on the video data from CCTV cameras provided by Osan Smart City Integrated Operations Center in Osan city, South Korea. Even though one of the origin purposes of these cameras is to record and deter street crime instances, some cameras deployed on crosswalks could be used to analyze potential risks such as estimation of PCRAs. We collected and processed the video footages from two crosswalks (Spots A and B) for 14 weekdays from January 9th to 28th during rush hours (from 8 am to 10 am and from 6 pm and 8 pm). Both spots are located near residential complexes and the floating population passing by is high. Spot A is designated as a school zone, but Spot B is not. In South Korea, school zones are certain roads near facilities for children under 13, including elementary schools, daycare centers and tutoring academies [22]. In such areas, if drivers cause accidents or break the rules in school zone, they will receive heavy penalties such as fines of up to 3,000 million won or imprisonment of life [22]. **Table 1** shows the actual CCTV views over crosswalks in Spots A and B, and describes the characters and meta-data of them.

In the preprocessing part, first, we partitioned the full video stream into "scenes (situations)" with motioned frames by using frame difference method, and traffic-related objects were detected and traced by using mask R-CNN model and deep SORT methods, respectively. In this step, we chose the scenes, as "interactive scenes" involved both vehicles and pedestrians in the scene at the same time.

As a result, we obtained about 2,122 and 1,915 trajectories for interactive scenes in Spots A and B, respectively. **Figure 3** shows the example trajectories after preprocessing.

**Table 1.** Information of the test spots

| | Spot A | Spot B |
|---|---|---|
| **CCTV view** | 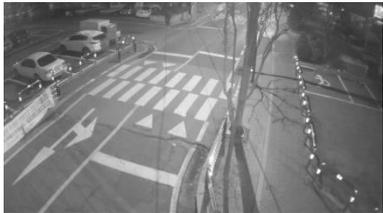 | 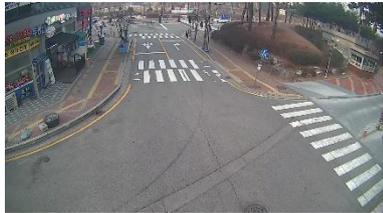 |

| Cam. name | Municipal Southern welfare/Daycare center #3 | Segyo complex #9 back gate #2 |
|---|---|---|
| **Crosswalk length (*m*)** | about 15 *m* | |
| **School zone** | ✓ | × |
| **# of lanes** | 2 | |
| **Signal light** | × | |
| **Speed limit (*km/h*)** | 30 *km/h* | |

**Note.** ✓: Yes,  ×: No,

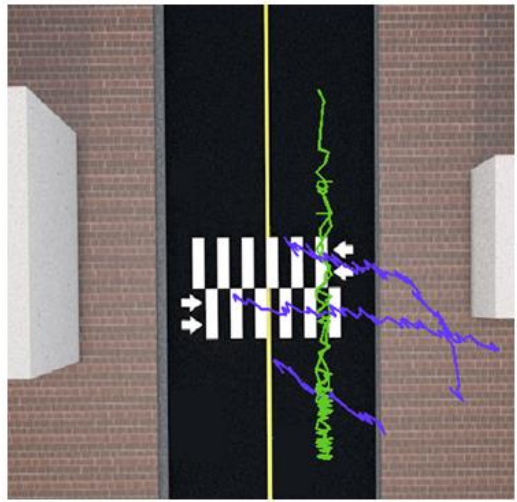

**Figure 3.** Examples of trajectories of vehicles and pedestrians after preprocessing

2.3. Trajectory prediction model based on deep LSTM networks

In this section, we describe the proposed trajectory prediction model based on deep LSTM networks, an evolution of recurrent neural network (RNN). The model trains the trajectories, and then predict trajectories with initial observed trajectories in test step (see **Figure 4**). In general, trajectories are time-series data, so RNN model can better handle the temporal dependencies of them. A core of this is to have chain-like structure of repeating processing and store information from previous processing steps. However, it is difficult for RNN model to process very long sequences because of gradient vanishing and exploding problems [23], [24].

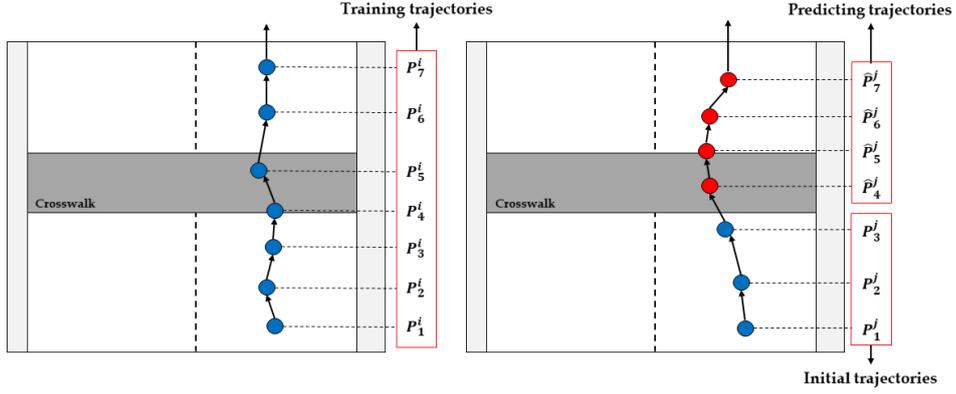

**Figure 4.** Concept of model training and test with trajectories

Unlike RNN structure, a key to LSTM network is a repeated chain with "cells", called memory blocks, $C_t$, with solving long-term dependency issue [25]. As described in **Figure 5**, typical LSTM network consists of three gates: (1) forget gate; (2) input gate; and (3) output gate [26], [27]. Forget gate decides what information forgetting adaptively by cell state [27]. In this gate, with values of $h_{t-1}$ and $x_t$, $f_t$ is calculated as follows:

$$f_t = \sigma(W_{xf} \cdot x_t + W_{hf} \cdot h_{t-1} + b_f)$$

where $\sigma(\cdot)$ is sigmoid function, $\cdot$ is elementwise product of the vectors, $h_{t-1}$ is hidden unit as output from previous LSTM layer, and $x_t$ is input from current input. $W_{*f}$ and $b_f$ are weight matrices and biases of forget gate, respectively. As a result, if this value is 0, previous information is completely removed, and if it is 1, previous one completely keeps.

Meanwhile, input and output gates control the flow of input activations into the memory cells and rest of network. Input gate decide whether the new information is stored, and to update the cell state with hyperbolic tangent function as follows:

$$i_t = \sigma(W_{xi} \cdot x_t + W_{hi} \cdot h_{t-1} + b_i)$$

$$\tilde{c}_t = tanh(W_{xc} \cdot x_t + W_{hc} \cdot h_{t-1} + b_c)$$

where $W_{*i}$ and $b_i$ are weight matrices and biases of input gate, respectively. Then, the new cell state is updated as follows:

$$c_t = f_t \cdot c_{t-1} + i_t \cdot \tilde{c}_t$$

The final step decides the current hidden unit, $h_t$ as input in next LSTM layer, based on output value as follows:

$$h_t = o_t \cdot tanh(c_t)$$

$$o_t = \sigma(W_{xo} \cdot x_t + W_{ho} \cdot h_{t-1} + b_o)$$

As a final result of output gate, we can obtain the estimated value $\hat{y}_t = \sigma(W_{hy}h_t + b_z)$ at time $t$. In order to minimize $\frac{1}{2}(y_t - \hat{y}_t)^2$ where $y_t$ is ground truth at time $t$, we derivate $z_t$ and $W_{hz}$ and error is backpropagated at the time step $t$.

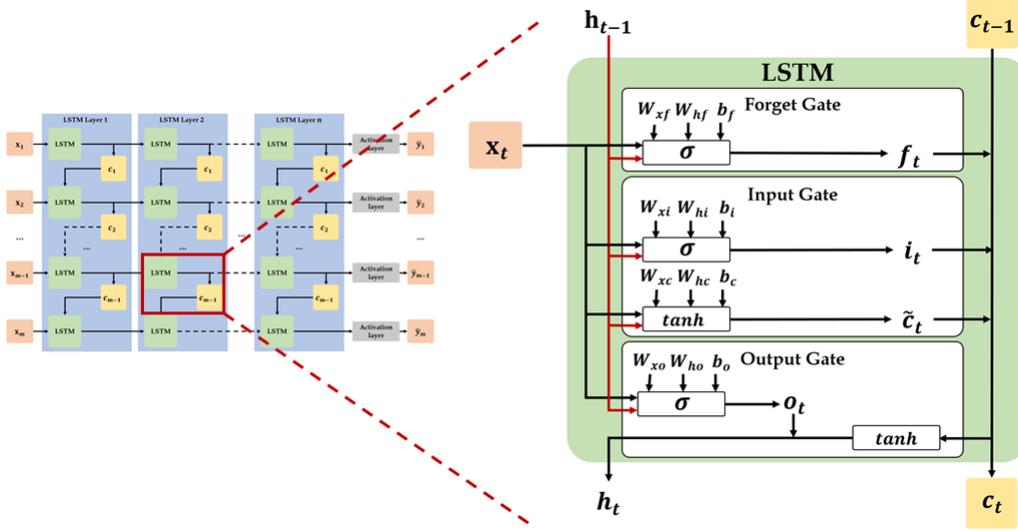

**Figure 5.** The structure of typical LSTM network consisting of forget, input, and output gates

In this study, we assume that the extracted trajectories of vehicle and pedestrian in consecutive timestamps are sequential data, and their trajectories can be predicted by using deep LSTM networks. In our experiment, one scene consists of trajectories for one vehicle and one pedestrian, so one scene has position information of each of them in consecutive timestamps $S_i = \{T_v^i, T_p^i\}$. In general, the trajectory of object, $T$, can be represented as pointwise by sequence as follows:

$$T^i = \{P_1^i, P_2^i, \dots, P_m^i\}; P_j^i = (x_j^i, y_j^i)$$

However, since the extracted trajectories in our experiment are from video, there is a slight error in each point. It means that most vehicles and pedestrians move in one direction in overhead viewpoint; from top to bottom and from left to right, but their trajectories can be shown as "zigzag" (see **Figure 3**). Thus, we first observed the "1 sec trajectories" of vehicles by shifting origin points of all trajectories to the same point, in which cars moved for a second, as illustrated in **Figure 6**.

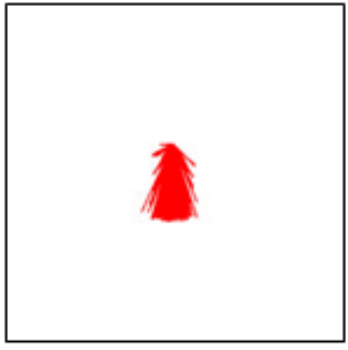

**Figure 6.** "1 sec trajectories" of vehicles shifted to the same origin point

As shown in **Figure 6**, trajectories are sector forms, so we anticipate that the estimated PCRA also has sector form, consisting of radius of circle and central angle. The trajectory $T$ can be represented as a sequence of velocity vector $\vec{v}_i$ between $P_i$ and $P_{i+1}$, each $\vec{v}_i$ is decomposed into speed ($|\vec{v}_i|$) and degree ($d_i$) with corresponding radius of circle and central angle, respectively. Thus, we can represent the trajectories, $T$, as follows:

$$T^i = \{\vec{v}_1^i, \vec{v}_2^i, \ldots, \vec{v}_{m-1}^i\}; \ \vec{v}_j^i = (|\vec{v}_j^i|, d_j^i)$$

From the predicted $\hat{\vec{v}}_i$, the predicted point can be calculated with simple trigonometric functions as follows:

$$P(\hat{x}_{i+1}, \hat{y}_{i+1}) = P(x_i + \Delta x, y_i + \Delta y)$$

$$\Delta x = \hat{\vec{v}}_i \times \sin\theta \ ; \Delta y = \hat{\vec{v}}_i \times \cos\theta\ , \ where\ \theta = \hat{d}_i \times \frac{\pi}{180°}$$

Thus, we construct trajectory prediction model consisting of two LSTM networks, called speed prediction network and degree prediction network (see **Figure 7**).

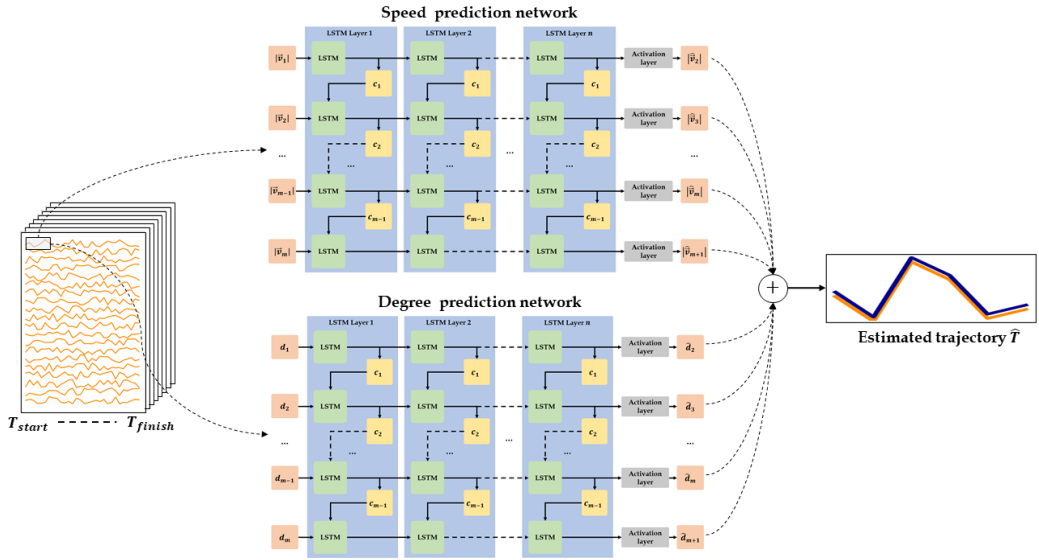

**Figure 7.** Trajectory prediction model using speed and degree prediction networks

In order to make the proposed model robust and estimate precise PCRAs, two issues have to be solved: (1) how to see the predicted trajectories far from the initially observed sequence, not an immediately next value (see **Figure 8 (a)**); and (2) how to handle the errors of predicted values in terms of speed and degree.

The first issue occurs by the structure of the LSTM model. It is difficult to see the predicted trajectories far from the initially observed sequence, not immediately next value. For example, assume that there is the observed sequence with three velocities $\{v_1, v_2, v_3\}$ (initial observed trajectory), then output sequence is $\{\hat{v}_2, \hat{v}_3, \hat{v}_4\}$, not $\{\hat{v}_4, \hat{v}_5, \hat{v}_6, \hat{v}_7\}$ we want. We can solve this issue by using the estimated values for the next estimation again, as illustrated **Figure 8 (b)**. In first round, from the given sequence $\{v_1, v_2, v_3\}$, we can obtain a sequence $\{\hat{v}_2, \hat{v}_3, \hat{v}_4\}$, and then make $\{v_2, v_3, \hat{v}_4\}$ as a new input sequence. Through the repetition of this procedure, we can obtain the final sequence $\{\hat{v}_4, \hat{v}_5, \hat{v}_6, \hat{v}_7\}$.

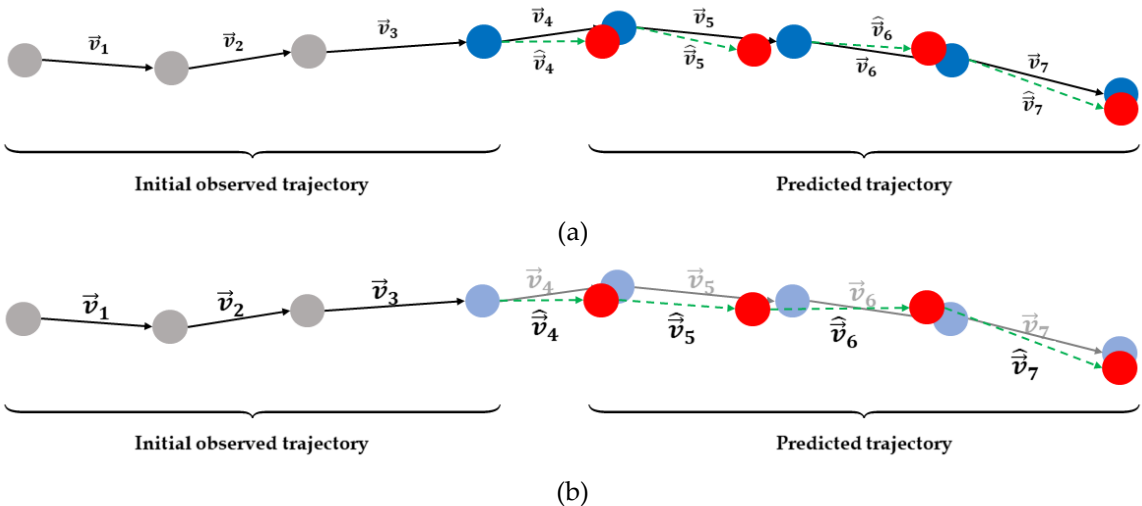

**Figure 8.** The methods of predicting future velocities by (a) one step; and (b) multiple steps

In the second issue, there are some residuals, $|y_t - \hat{y}_t|$, between the observed velocities and predicted velocities. Thus, in order to handle such errors, we infer the confidence intervals of speed and degree with the given trajectories, and then PCRAs are estimated. The detailed procedure will be described in the next section.

2.4. Estimation of predictive collision risk area using statistical inference

In this section, we explain how to infer the PCRAs based on the predicted trajectories. In practice, it is difficult for the proposed prediction model to accurately predict the future trajectories, and there are some tolerances. These tolerances of the model can be presented as confidence intervals, which have the ability to quantify the uncertainty of the estimate. It provides both lower and upper limits and a likelihood. In this paper, we assume that the predicted speed and degree in point $\hat{P}_{i+1}$ are mean values of their distributions. Consequently, the confidence intervals consist of speed lower and upper limits (bounds) and degree lower and upper limits (bounds), and in graphically, the shape is a truncated sector in concentric circles as illustrated in **Figure 9**. In this paper, we define "PCRA" in form of sector as a space enclosed by the boundaries of degree lower and upper bounds, and speed upper bound, as described in **Figure 9**. Since the vehicle moves continuously, we do not consider speed lower bound.

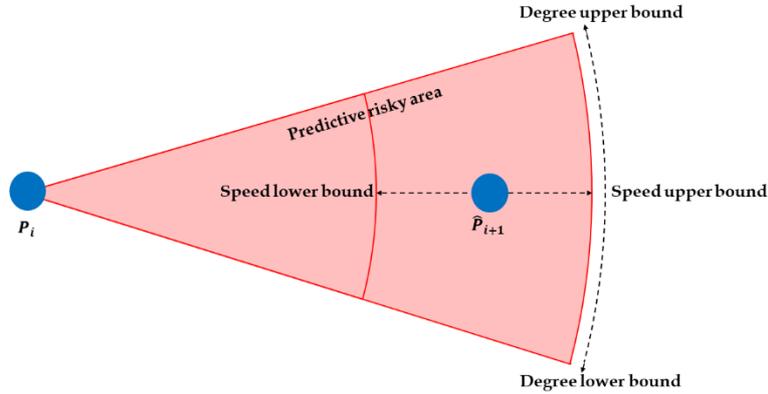

**Figure 9.** Confidence intervals of speed/degree, and PCRA

In order to find confidence boundaries, we need to figure out probability distribution functions (PDFs) of speed and degree considering the zone number and future time second. As seen in **Figure 10**, all objects would pass certain zones in RoI (Region of Interest), and we can obtain the speed and degree information after $t_j$ time second (e.g., 1 sec. and 2 sec.) when they pass by this zone. In addition, since the objects have different velocities according to zones, we split the RoI into $N$ zones. Thus, the PDFs are denoted as $f_{v,z_i,t_j}(x)$ and $f_{d,z_i,t_j}(x)$, respectively. $f_{v,z_i,t_j}(x)$ means probability distribution function of speed/degree after $t_j$ in zone $z_i$. For example, $f_{v,1,2}(x)$ means the PDF of speed after 2 seconds in zone number 1.

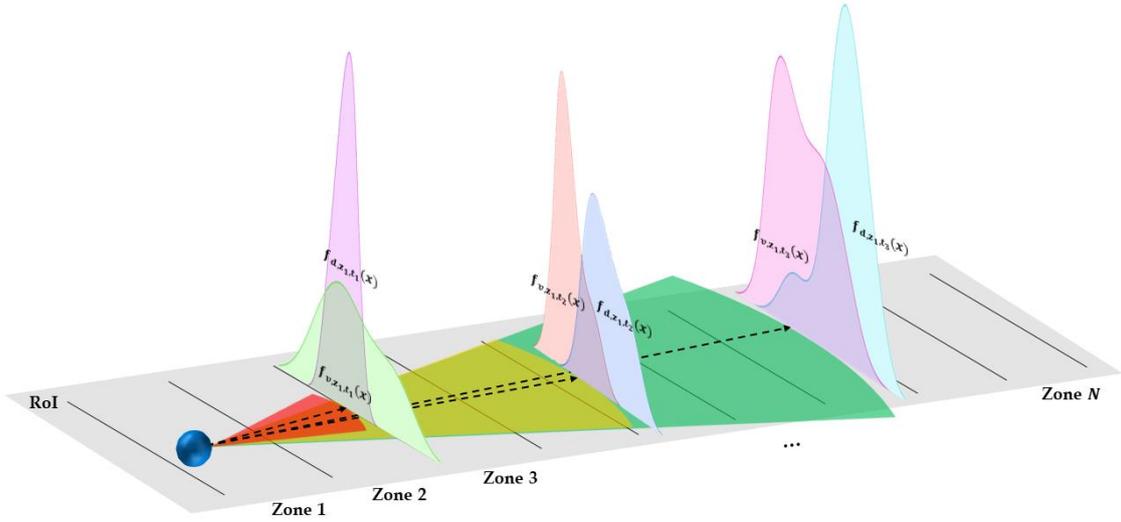

**Figure 10.** Example of PDFs of sped and degree in zone 1 after $t_1$, $t_2$ and $t_3$ sec.

With the obtained PDFs, the confidence intervals can be calculated. Our first strategy is to get the confidence intervals with mean and standard deviation if the collected data are fitted with a normal distribution. First, we conducted normality test (Shaprio-Wilks test [28], [29]) in order to determine whether the given data follows a normal distribution with following hypothesis:

Null hypothesis $H_0$: Data follows a normal distribution
Alternative hypothesis $H_1$: Data do not follow a normal distribution

However, the given data do not follow normal distribution, so $H_0$ is rejected (p-value $\leq 0.05$). In addition, it does not follow any probability distributions. Instead, we adopted non-parametric statistical inference strategy using empirical cumulative distribution function (ECDF) [30], [31]. Typically, ECDF provides a simple estimate of an unknown probability distribution (distribution-free) by "best-guessing" of the true form of the underlying unknown distribution $F(x)$. Given a dataset $x_1, x_2, \ldots, x_n$ of independent, identically distributed random observation, the ECDF is obtained as follows:

$$\hat{F}(x) = \frac{\text{\# of elementes in the sample} \leq x}{n} = \frac{1}{n}\sum_{i=1}^{n} 1_{x_i \leq x}$$

where $1_A$ means an indicator of event $A$. The ECDF is non-decreasing function between 0 and 1, and $m$ observation taking value $x_i$ causes the empirical distribution function to increase by amount $m/n$ at $x_i$. In general, $\hat{F}(x)$ converges to $F(x)$ almost surely for every $x$ values when $n$ approaches to infinite [32].

Meanwhile, confidence intervals in ECDF are defined as a function with fixed $x$ value and $1 - \alpha$ pointwise confidence interval about $x$ as follows [33]:

$$P(\{F(x) \in C(x)\}) \geq 1 - \alpha$$

where $C(x)$ is the region by the pointwise confidence interval. Our issue is that while each $x$ value obeys the confidence interval because pointwise confidence interval is computed for each $x$ value, it does not guarantee $C(x)$ would obey $1 - \alpha$ confidence region for each individual $x$ value at the same time [34]. Instead, we can define $1 - \alpha$ confidence band for all $x$ as follows:

$$P(\{F(x) \in B(x), \forall x\}) \geq 1 - \alpha$$

where $B(x)$ is function for confidence band. Then, by Dvoretzky-Kiefer-Wolfowitz (DKW) inequality, we can obtain two-sided confidence bands (called lower bound $L(x)$, and upper bound $U(x)$) as follows [35]:

$$P\left(\sup_{x \in \mathbb{R}} |\hat{F}(x) - F(x)| > \varepsilon \right) \leq \alpha = 2e^{-2n\varepsilon^2}$$

$$L(x) = max\{\hat{F}(x) - \varepsilon,\ 0\}$$
$$U(x) = min\{\hat{F}(x) + \varepsilon,\ 1\}$$

where $\varepsilon = \sqrt{(\log 2/\alpha)/2n}$, and **Figure 11** shows an example of ECDF and lower and upper confidence bands.

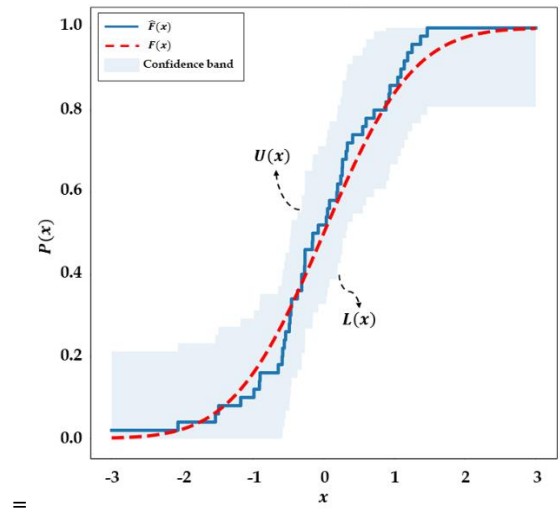

**Figure 11.** Example of $\hat{F}(x)$, $F(x)$, and confidence bands

In our experiment, we calculate ECDFs and their upper and lower confidence bands, $L(x)$ and $U(x)$ in all zones for future time second $t_j$ in order to obtain PCRAs.

## 3. Experiments and Results

### 3.1. Experimental design

In this section, we briefly describe the datasets, and design trajectory prediction models and ECDFs to figure out the severity of potential risks. Before conducting the main experiment, we smoothed the trajectories (for speeds and degrees in each scene). In general, since the data extracted from video have noise, it needs to smooth the filtering algorithm such as low-pass filter [36].

In our experiment, we conduct mainly two experiments, and the goals of them: (1) to find the optimal trajectory prediction models; and (2) to infer the PCRA and classify the severity of potential risks.

Trajectory prediction model design: Note that the proposed trajectory prediction model consists of two networks; speed prediction network and degree prediction network. In addition, we designed two models for each target spot because the trajectory patterns are different in each spot (see **Figure 12**). Since the typical performance deep LSTM network depends on various factors such as numbers of LSTM layers (hidden layers), numbers of cells as well as other hyperparameters (dropout ratio, optimizer type, etc.).

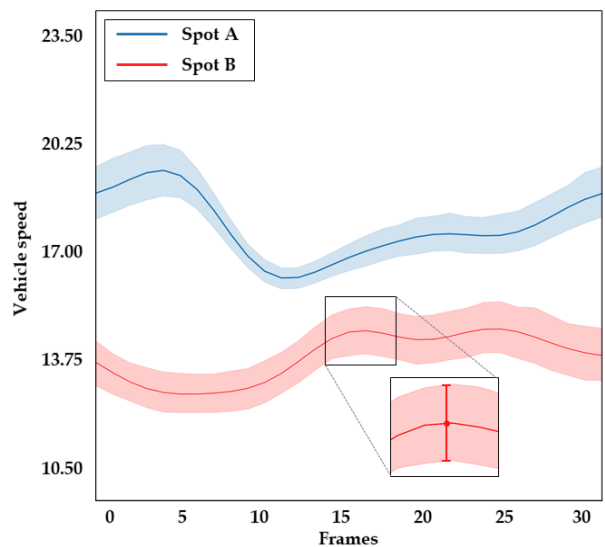

**Figure 12.** Average speed of vehicles by frames in each test spot

Thus, we design 10 models by the numbers of hidden layers and cells as described in **Table 2**. These models are applied to predict vehicle speed and degree, and pedestrian speed and degree in Spots D and G (to find total 6 models). The number of training and test data are about 70% (Spot A: 1,478, Spot B: 1,294) and 30% (Spot A: 644, Spot B: 621) of all trajectories in each spot.

Table 2. Designed model structures and learning hyperparameters

| Model name | # of hidden layers | # of cells | Dropout ratio | Optimizer | Loss function | Sliding window size | Epoch |
|---|---|---|---|---|---|---|---|
| Simple-TP | 1 | 3 | 0.2 | Adaptive moment estimation (ADAM) | Mean squared error (MSE) | 3 | 80 |
| Deep-TP-3-10 | 3 | 10 | | | | | |
| Deep-TP-3-40 | 3 | 40 | | | | | |
| DeepTP-3-80 | 3 | 80 | | | | | |
| Deep-TP-5-10 | 5 | 10 | | | | | |
| Deep-TP-5-40 | 5 | 40 | | | | | |
| Deep-TP-5-80 | 5 | 80 | | | | | |
| Deep-TP-10-10 | 10 | 10 | | | | | |
| Deep-TP-10-40 | 10 | 40 | | | | | |
| Deep-TP-10-80 | 10 | 80 | | | | | |

Criteria for predictive collision risk area: In order to conduct PCRA inference and measure potential risk, we need to set zone number $z_i$ and future time $t_j$ in $f_{*,z_i,t_j}(x)$. First, the RoI in target spots is split into 12 zones for vehicle (horizontal) and pedestrian (vertical). In our experiment, we set $t_j$ as 1, 2, and 3 (after 1 sec, 2 sec, and 3 sec) based on PIEV (perception-intellection-emotion-volition) theory in accident [37]–[40]. PIEV is the amount of time it takes a driver to react to a hazard, and each step is time to discern an object or event, to understand the implications of the situation, to decide how to react, and to initiate the action (engaging the brakes), respectively. Much research has been conducted on PIEV time, and designed various time ranges for safe, 1.48 to 2.5 seconds [38]–[40]. ASSHTO (The America Association State Highway and Transportation Officials) recommended to secure 2 till 2.5 seconds [37]. Thus, the prototype of the proposed system can adjust the time divisions for severity of potential risk depending on the road environment. In our experiment, we categorized severity as danger, warning, and relative safe, and defined them by overlapped areas as follows:

- Danger: When areas of $t_j = 1$ for vehicle and pedestrian are overlapped.
- Warning: When areas of $t_j = 2$ for vehicle and pedestrian are overlapped.
- Relative safe: When areas of $t_j = 3$ for vehicle and pedestrian are overlapped, or not overlapped.

When determining severity of potential risks, the danger case always takes precedence over warning and relative safe cases. **Figure 13** shows the warning case for the overlapped areas for vehicle (blue point) and pedestrian (green point).

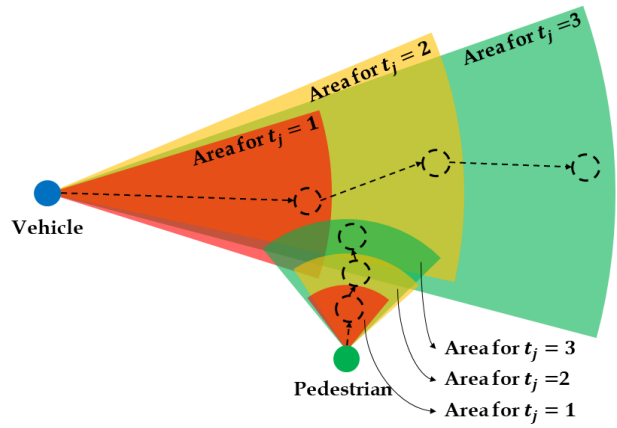

**Figure 13.** Risk cases depending on the overlapped areas for vehicle and pedestrian

3.2. Validation of trajectory prediction model

In this section, we validate the designed models, and choose the best performance models for trajectory prediction of vehicle and pedestrian. First, **Figure X14 (a)** to **(h)** illustrate the results of loss values by epoch for vehicle/pedestrian's speed/degree in Spots A and B.

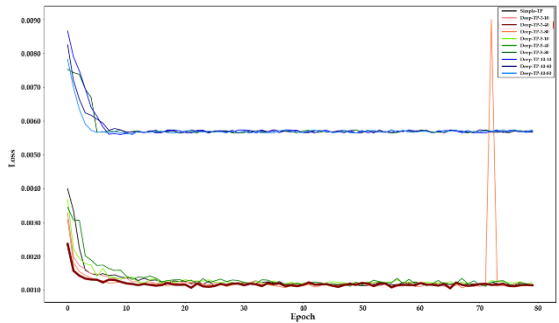

(a) For vehicle speed in Spot A

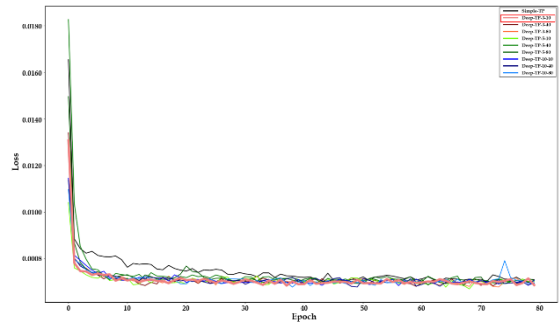

(b) For vehicle degree in Spot A

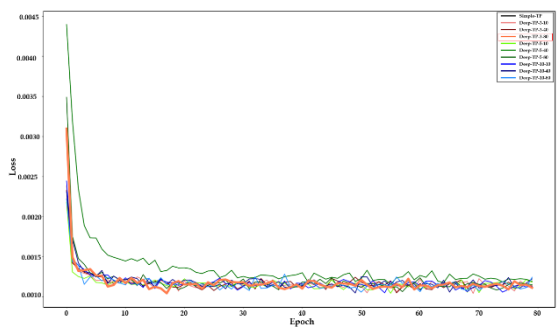

(c) For vehicle speed in Spot B

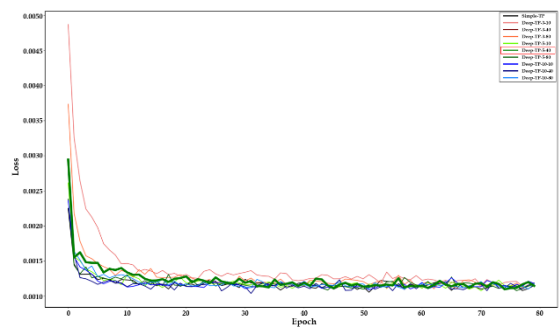

(d) For vehicle degree in Spot B

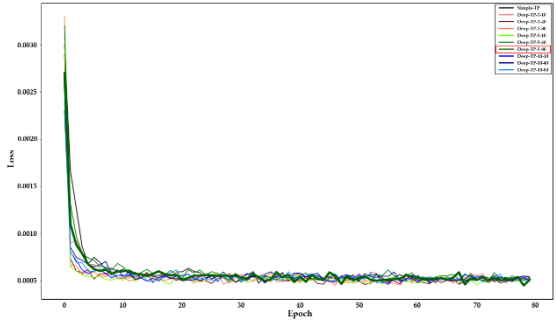
(e) For pedestrian speed in Spot A

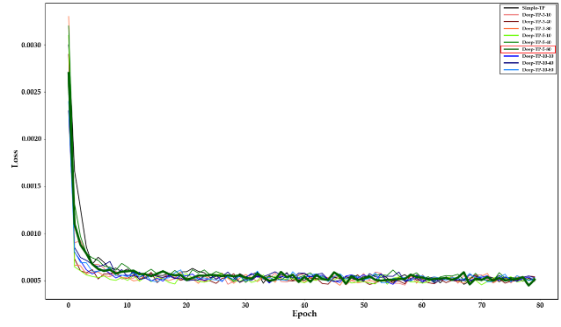
(f) For pedestrian degree in Spot A

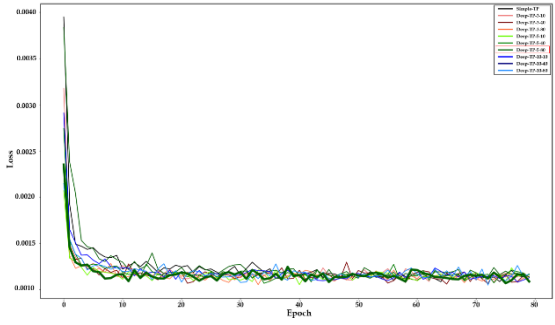
(g) For pedestrian speed in Spot B

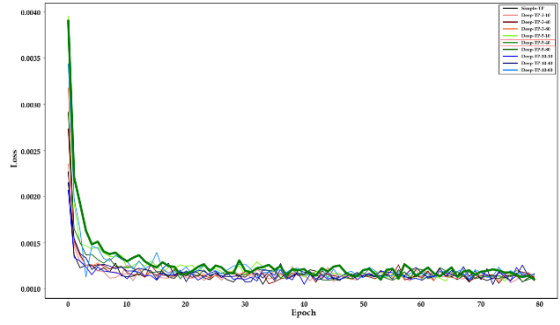
(h) For pedestrian degree in Spot B

Figure 14. Training results of the designed models

Most models make loss values reduce as iterating learning. In **Figure X14 (a)**, Deep-TP-3-80 model has suddenly jumped value (maybe overfitted), and when the size of the model grew, the training loss converged in the local minimum, not reduced.

**Table 3** shows the best training performance models and MSE tested by these models. In general, the test performance of sequence data is measured by MSE, and values for preceding periods are input, and the values for remaining subsequent periods are compared to the predicted values, as illustrated in **Figure 15 (a)**. However, in our experiment, one input data is one trajectory sequence (e.g., four subsequent velocities are expected by eight preceding velocity), so MSE is calculated in trajectory level as represented in **Figure 15 (b)**. With the expected velocities, the PCRAs are inferred.

**Table 3.** Best performance models for each dataset and test results

| Target spot | Target dataset | Best training performance model | Test MSE |
| --- | --- | --- | --- |
| Spot A | Vehicle speed | Deep-TP-3-40 | 0.0289 |
| | Vehicle degree | Deep-TP-3-10 | 0.0310 |
| | Pedestrian speed | Deep-TP-5-80 | 0.0105 |
| | Pedestrian degree | Deep-TP-5-40 | 0.0128 |
| Spot B | Vehicle speed | Deep-TP-3-80 | 0.0221 |
| | Vehicle degree | Deep-TP-5-40 | 0.0293 |
| | Pedestrian speed | Deep-TP-5-80 | 0.0098 |
| | Pedestrian degree | Deep-TP-5-40 | 0.0129 |

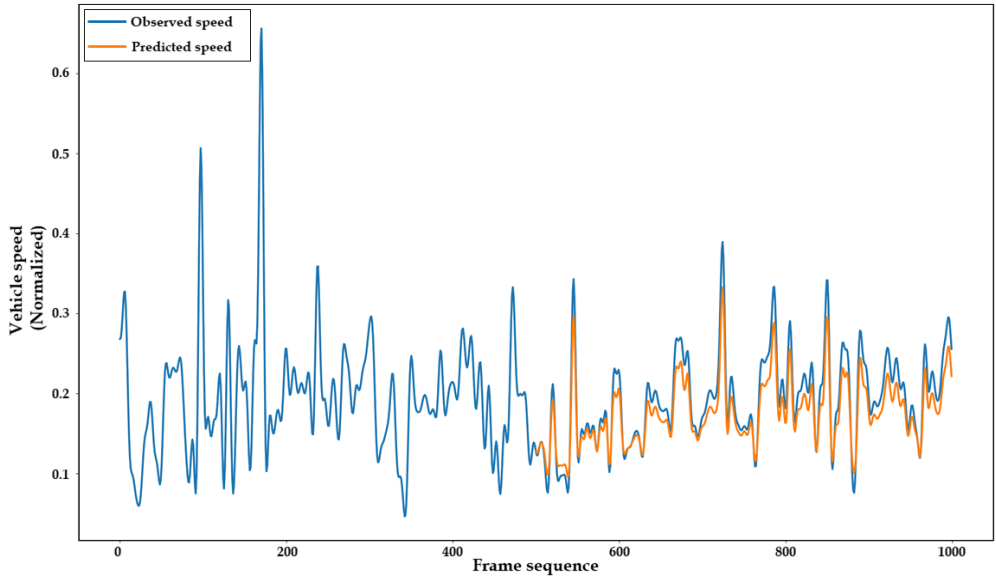
(a) Typical prediction processing for preceding period and remaining subsequent period

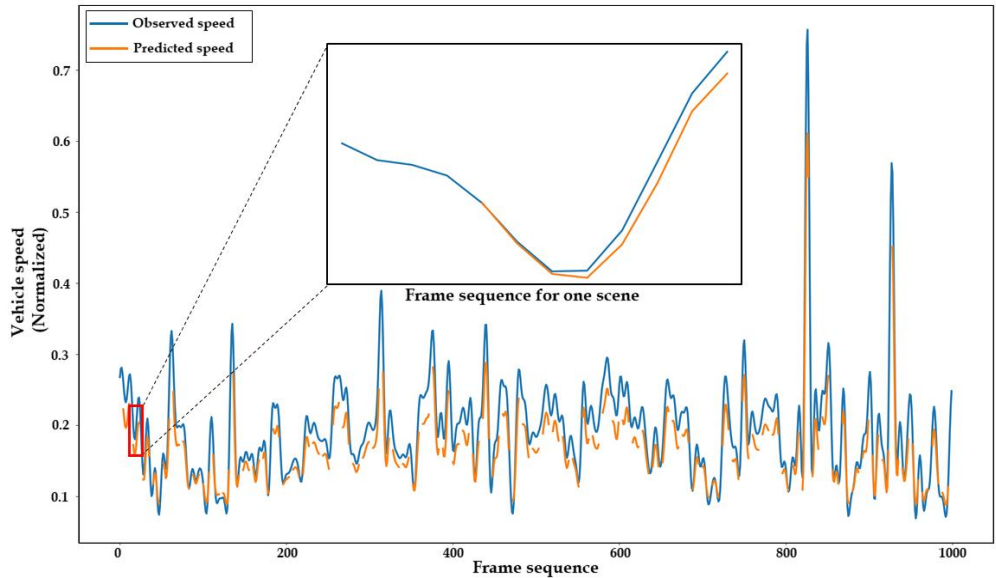
(b) Trajectory level prediction in total sequence
**Figure 15.** Model evaluation strategies

3.3. Measurement of PCRA in testbed

In this section, we infer PCRAs, and measure the severity of potential risks in test spots. Note that the predictive potential risky area is based on predicted trajectory (velocity) and statistical inference using ECDF. In our experiment, we set the α as 0.05 (95% confidence interval). **Figure 16 (a)** and **(b)** describe the sample ECDF plots for pedestrians' speed and degree. From these figures, we can know that distribution of pedestrian speed is concentrated in a certain range (where y-values increase rapidly), and distribution of pedestrian degree is spread across various values. It means that most pedestrians move at a similar speed, and their walking directions are wide. In practice, all ECDFs of time and zone are applied.

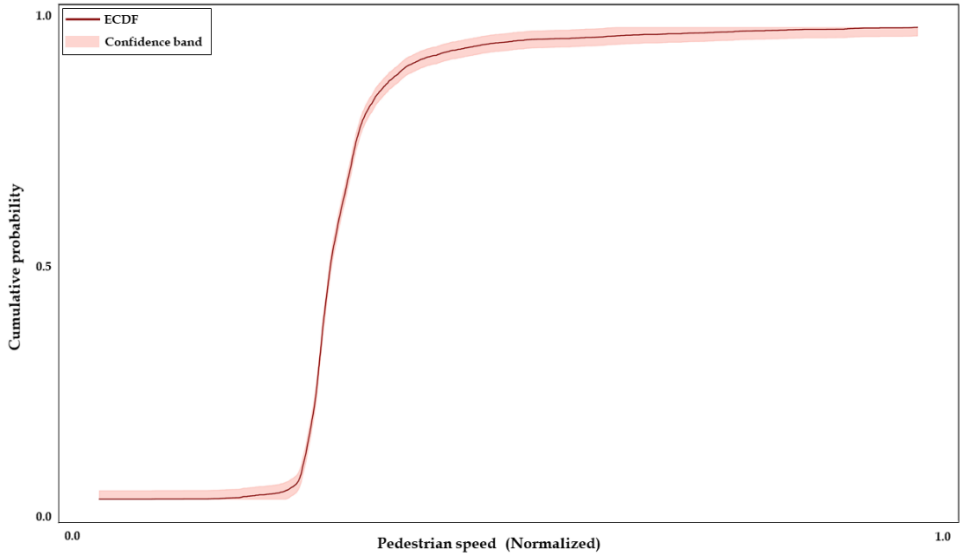
(a) ECDF for pedestrian speed for after 1 sec. in zone 1 in Spot A

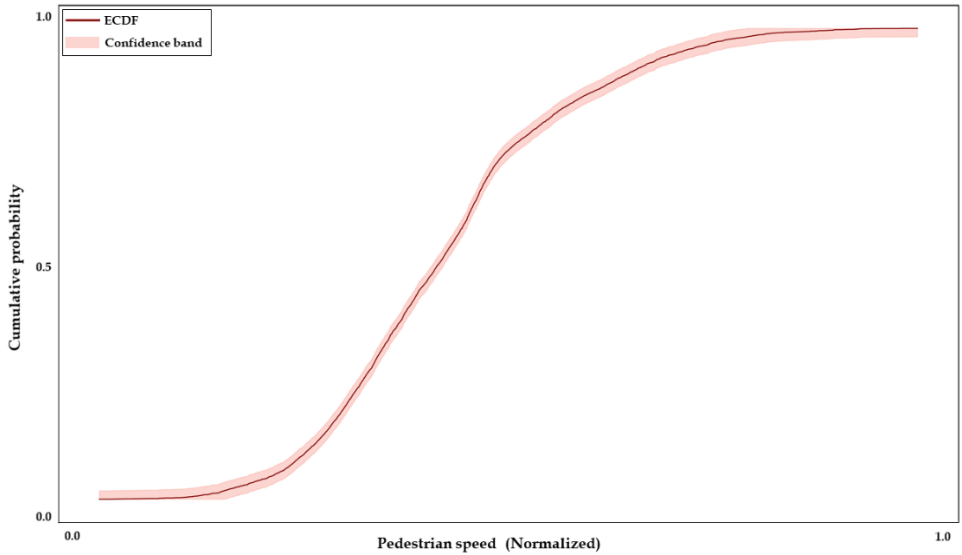
(b) ECDF for pedestrian degree for after 1 sec. in zone 1 in Spot B

**Figure 16.** Sample ECDF plots for pedestrian speed and degree

Based on overlap of PCRAs between vehicle and pedestrian, we can measure the potential risks in each spot. The results are described in **Table 4**. As a result, the dangerous scenes occurred more in Spot A than in Spot B. The ratio of warning scenes is similar in both locations (0.043 and 0.035, respectively), but the ratio of dangerous scenes is higher in Spot A (0.115) than in Spot B (0.077). It means that Spot A would be more dangerous than Spot B. In the road environmental perspective, Spot A has fences separating the road from sidewalk and the raised crosswalk (bumper-typed crosswalk), so it seems safer. However, our experimental result shows that Spot A is a more dangerous location. As we guess, since surrounding obstacles narrow a pedestrian's view in this spot, they would be on the road to check safe margins before crossing. On the other hand, Spot B has no obstructions to pedestrian's vision, so it is possible to recognize the approaching vehicle from farther away. Finally, we implemented a prototype to assess the predictive potential risks. Final result video is uploaded in Youtube channel [41], and **Figure 17** is a snapshot of this video.

**Table 4.** Results of measuring potential risks in each spot using the proposed system

| Target spot | # of total scenes (for test data) | Results | |
|---|---|---|---|
| | | # of dangerous scenes | # of warning scenes |
| Spot A | 644 | 74 | 28 |
| Spot B | 621 | 48 | 22 |

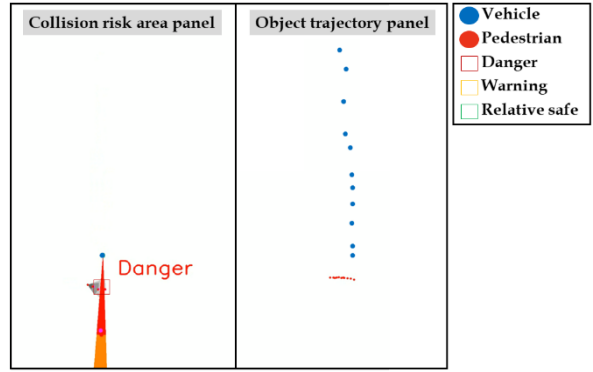

**Figure 17.** Snapshot of the result

3.4. Discussions

In this paper, we proposed a predictive collision risk area estimation system using deep LSTM networks and statistical inference as part of the overall architecture for proactive pedestrian accident prevention based on vision. This part mainly consists of two modules; object's trajectory prediction using deep LSTM networks and PCRA estimation using statistical inference. There are three contributions in this research: (1) to apply the vision-based data to recognize potential risks for vehicle-pedestrian interactions; (2) to predict trajectories of traffic-related objects and infer their predictive boundaries; and (3) to divide the severity of potential risk quantitatively depending on the overlapped areas between vehicle's and pedestrian's PCRAs.

In general, the behaviors of traffic-related objects have different patterns depending on the road environment such as fence-deployed road, unsignalized crosswalk, and school zone [21]. Thus, we designed the prediction networks and trained trajectories by test spots. Much research has been conducted on forecasting object's trajectory using social graph convolutional LSTM [42], spatio-temporal graph transformer networks [43], probabilistic crowd GAN [44] based on deep learning methods. These studies handled only one class's trajectory forecast, but our study aims to predict trajectory in interactive situations between vehicle and pedestrian. In addition, our system considers the confidence interval for predicted positions, so it can provide the driver/pedestrian with the deliberate information of threat. **Table 5** compares the existing studies for potential risks between vehicle and pedestrian with our approach. To measure potential risk, these systems applied various standardized features such as TTC and PET with predicting trajectories, but not providing risk severity division. The closest approach to ours was [45], which similarly targeted near-miss collisions between vehicle and pedestrian based on video sources along with survey. By contrast, our process used only video data, and further can categorized severity of levels as three divisions; danger, warning, and relative safer according to time second, $t_j$. In our experiment, we set this value conservatively (1 sec, 2 sec, and 3 sec), but it can be adjusted by decision-makers or field administrators depending on the road environment or weather.

**Table 5.** Comparison table for characteristics of previous systems with our approach

|  | Data source | Target spot | Target behavior | Risk measurement | Trajectory prediction method | Severity estimation method | Risk severity |
|---|---|---|---|---|---|---|---|
| Ref [45] | Video (tilted view), and videographic survey | Varying road geometry (midblock crosswalks) | Near-miss collision (V-P) | PET, conflict rate | × | PET-based severity levels | + (Numerical values) |
| Ref [46] | Video (onboard view) | Various roads for pedestrians to cross | Near-miss collision (V-P) | TTC | × | × | n/a |
| Ref [47] | Video (tilted view) | Unsignalized intersection | Vehicle's turning | × | LSTM networks | × | n/a |
| Ref [15] | Video (top view) | Unsignalized crosswalk | Near-miss collision (V-P) | × | W/CDM-MSFM | × | n/a |
| Ref [14] | Video (tilted view) | Signalized intersection | Red-light crossing intentions | PET | LSTM networks | × | n/a |
| Ref [48] | Video (tilted view) | Signalized intersection | Near-miss collision (V-P) | Crossing intention | SVM and SLP | × | n/a |
| Proposed system | Video (tilted view) | Unsignalized crosswalk | Near-miss collision (V-P) | Potential collision risk areas | LSTM networks | Empirical cumulative density function using speed and degree | + (Danger, warning, relative safe) |

**Note.** +: Provided ×: No provided, n/a: Not applicable, TTC: Time-to-collision, PET: Post encroachment time, SVM: Support vector machine, SLP: Simple layer perceptron, V-P: Vehicle-pedestrian

It should be noted that the proposed system was motivated by a lack of an efficient potential risk recognition and its severity division quantitatively, in order to prevent pedestrian accidents proactively. Further, we believe that this can be a surrogate measurement evaluating vehicle-pedestrian collision risk, and provide warning with drivers/pedestrians in bi-directions as proliferation of V2X and P2X communications.

## 4. Conclusions

In this study, we proposed a predictive collision risk area estimation system using deep LSTM model and statistical inference. The cores of the methodologies are deep LSTM networks using speed and degree and ECDF models in order to predict traffic-related object's trajectory and estimate predictive collision risk area, respectively. In addition, we defined measurement for severity of potential risks by overlapped areas between vehicle and pedestrian, as danger, warning, and relative safe. In our experiment, we assessed the levels of potential risks in Spots A and B. First, in order to predict trajectories, we chose the proposed networks among the designed models depending on their structures. Next, with the predicted trajectories, the predictive potential risky areas were estimated by using the ECDF and inferring the confidence interval with zone and future time parameters. As a result, contrary to our expectations, Spot A is more dangerous than Spot B because the obstacles narrow the pedestrian's vision.

We validated the feasibility of our proposed system by implementing the prototype and applying it to actual spots in Osan city, South Korea. The proposed system can provide information about predictive warning with drivers/pedestrians, and makes it possible to take actions to prevent potential collisions. With the measuring such risks, in the future, decision-makers, such as urban planners and safety administrators, can apply and analyze them to gain a better understanding of where near-miss collision situations frequently occur. In addition, they collaborate using these clues to improve the safety of the spaces by deploying safe structures, such as speed cameras and bumpers in order to prevent collisions proactively, and this is a part of our ongoing work.